\documentclass[10pt,twocolumn,letterpaper]{article}

\usepackage{cvpr}
\usepackage{times}
\usepackage{epsfig}
\usepackage{graphicx}
\usepackage{amsmath}
\usepackage{amssymb}
\usepackage{gensymb}
\usepackage{xcolor}
\usepackage{caption}
\usepackage{subcaption}
\usepackage{textcomp}
\usepackage{float}
\usepackage{dirtytalk}
\usepackage{breqn}

\usepackage[pagebackref=true,breaklinks=true,letterpaper=true,colorlinks,bookmarks=false]{hyperref}

\cvprfinalcopy 

\ifcvprfinal\fi

\makeatletter
\def\blfootnote{\xdef\@thefnmark{}\@footnotetext}
\makeatother

\begin{document}

\title{Radar Camera Fusion via Representation Learning in Autonomous Driving}

\author{Xu Dong \qquad Binnan Zhuang  \qquad Yunxiang Mao \qquad Langechuan Liu \textsuperscript{$\dagger$} \\
XSense.ai\\
11010 Roselle Street, San Diego, CA 92121\\
}

\maketitle
\thispagestyle{empty}

\blfootnote{\textsuperscript{$\dagger$} {indicates corresponding author \tt patrickl@xsense.ai}}
\blfootnote{Note: this paper was published on CVPR 2021.}

\begin{abstract}
Radars and cameras are mature, cost-effective, and robust sensors and have been widely used in the perception stack of mass-produced autonomous driving systems. Due to their complementary properties, outputs from radar detection (radar pins) and camera perception (2D bounding boxes) are usually fused to generate the best perception results. The key to successful radar-camera fusion is the accurate data association. The challenges in the radar-camera association can be attributed to the complexity of driving scenes, the noisy and sparse nature of radar measurements, and the depth ambiguity from 2D bounding boxes. Traditional rule-based association methods are susceptible to performance degradation in challenging scenarios and failure in corner cases. In this study, we propose to address radar-camera association via deep representation learning, to explore feature-level interaction and global reasoning. Additionally, we design a loss sampling mechanism and an innovative ordinal loss to overcome the difficulty of imperfect labeling and to enforce critical human-like reasoning. Despite being trained with noisy labels generated by a rule-based algorithm, our proposed method achieves a performance of 92.2\% F1 score, which is 11.6\% higher than the rule-based teacher. Moreover, this data-driven method also lends itself to continuous improvement via corner case mining.

\end{abstract}

\section{Introduction}
LiDAR, radar, and camera are the three main sensory modalities employed by the perception system of an autonomous driving vehicle. Though LiDAR-based 3D object detection is very popular in high-level autonomy, its wide adoption is still limited by some unsolved issues. First, LiDAR is prone to adversarial conditions (e.g. rainy weather); second, current LiDAR systems still exhibit prohibitively high maintenance need and cost; third, the mass-production of LiDAR is not ready to meet the growing demand. 

\begin{figure}[t]
\begin{center}
  \includegraphics[width=1.0\linewidth]{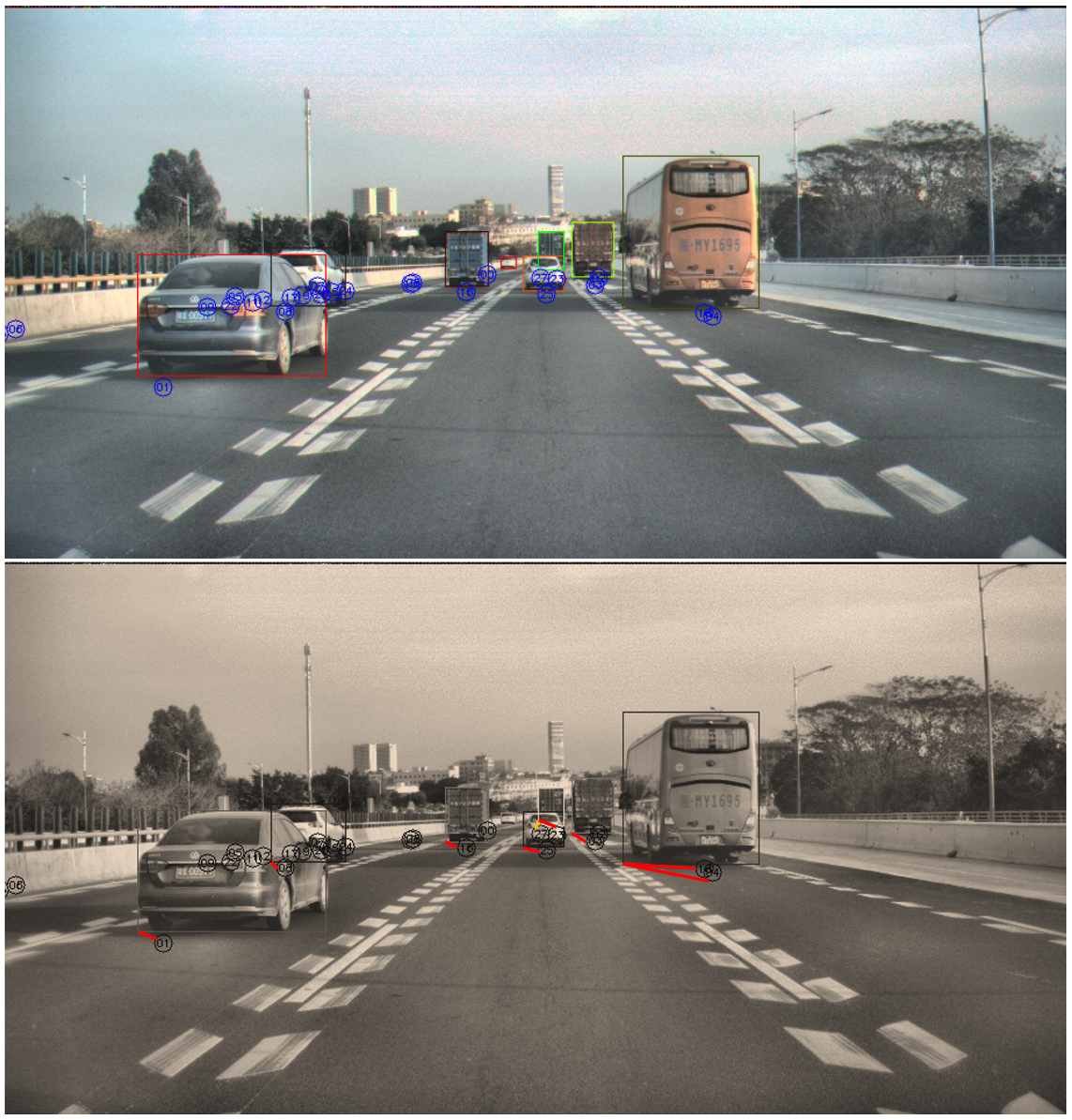}
\end{center}
\setlength{\abovecaptionskip}{7pt} 
  \caption[]{An illustration of the associations between radar detections (radar pins) and camera detections (2D bounding boxes). The context of the scene is illustrated in the top picture, with the image captured by the camera along with the detected bounding boxes and the projected radar pins (shown as numbered blue circles). The bottom picture adds red lines to highlight the association relationships between radar pins and bounding boxes. The tiny orange line in the middle denotes \textit{uncertain} association relationship, which will be explained later.}
\label{fig:showcase}
\end{figure}

\begin{figure*}[h]
\begin{center}
  \includegraphics[width=0.95\linewidth]{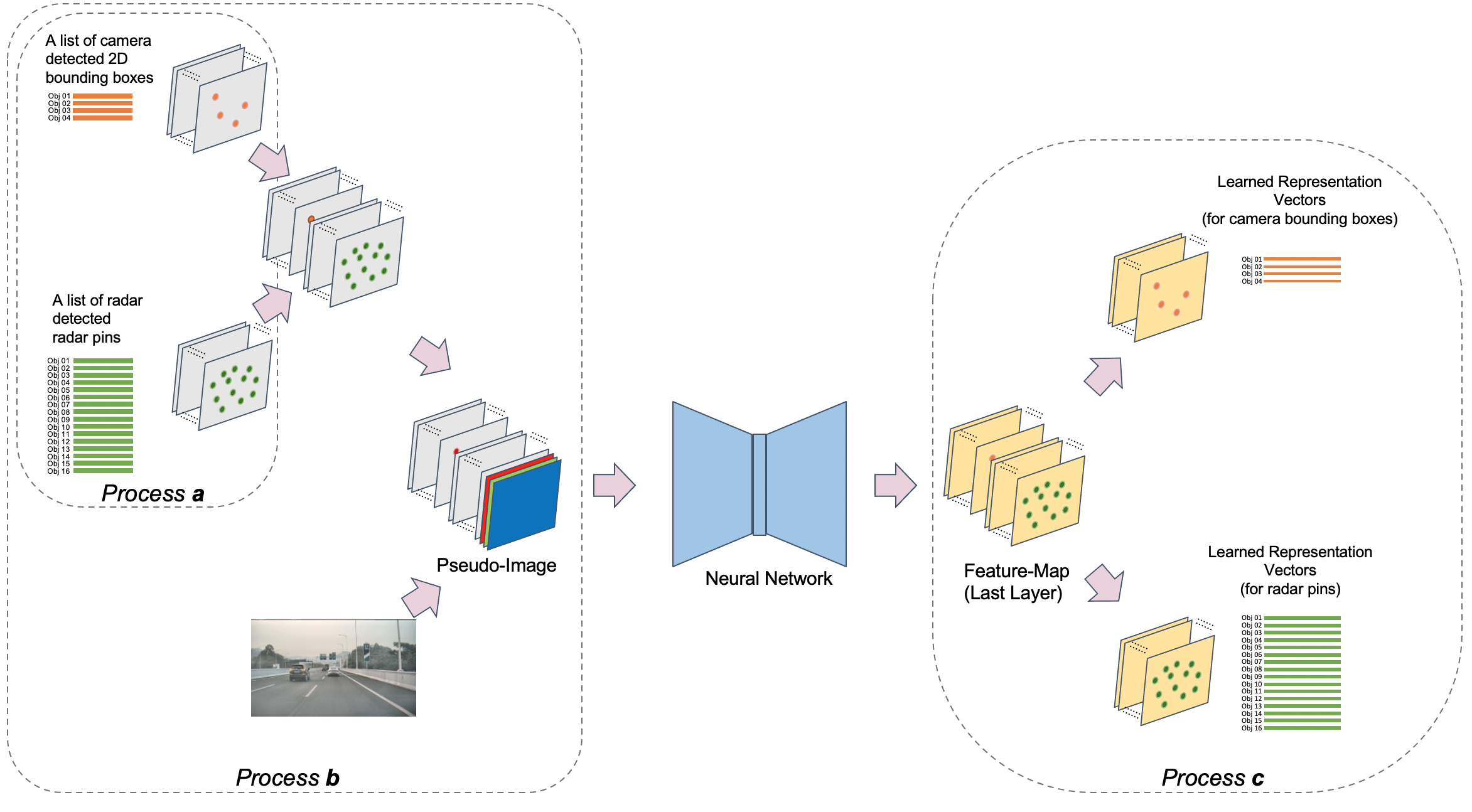}
\end{center}
\setlength{\abovecaptionskip}{4pt} 
  \caption{An overview of AssociationNet. \textit{Process \textbf{a}} illustrates how the radar pins and 2D bounding boxes are first projected into the camera image plane and then produce a pseudo-image. \textit{Process \textbf{b}} illustrates how the final pseudo-image is composed by concatenating all the features of radar pins, bounding boxes, and the original RGB camera image. The pseudo-image is then fed into a neural network to learn high-level semantic representations. \textit{Process \textbf{c}} illustrates how the learned representation vectors for objects are finally extracted from the feature-map generated in the last layer of the neural network.}
\label{fig:overview}
\end{figure*}

An automotive millimeter-wave radar can also provide a certain level of geometrical information with relatively precise range and speed estimates. Moreover, as a widely-adopted sensor in automobiles for decades, radar is relatively robust, low-cost, and low-maintenance. The fusion between radar and camera combines radar's geometrical information and camera's appearance and semantic information, which is still the mainstream perception solution in many practical autonomous driving and assisted driving systems.

Traditionally, the radar-camera fusion is achieved by the combination of rule-based association algorithms and kinematic model-based tracking. The key is data association between radar and camera detections. The noisy and sparse nature of radar detection and the depth ambiguity from a mono camera makes such association problem very challenging. Traditionally, the association process is hand-crafted based on minimizing certain distance metrics along with some heuristic rules. It not only requires a large amount of engineering and tuning but is also hard to adapt to ever-growing data.

An emerging solution is to use learning-based methods to replace the rule-based radar-camera fusion. The latest advances focus on direct 3D object detection with the combined radar and camera data as the input \cite{john2019rvnet, nabati2019rrpn, nobis2019deep}. These approaches all rely on LiDAR-based ground-truth to build the link between radar and camera. This is feasible on most public datasets such as nuScenes~\cite{nuscenes2019}, Waymo~\cite{sun2019scalability} etc. However, it cannot be applied to a large fleet of commercial autonomous vehicles, often equipped with only radars and cameras. In this study, we propose a scalable learning-based framework to associate radar and camera information without the costly LiDAR-based ground-truth.

Our goal is to find representations of radar and camera detection results, such that matched pairs are close and unmatched ones are far. We convert the detection results into image channels and combine them with the original image to feed into a convolutional neural network (CNN), namely, \textit{AssociationNet}. Training is performed based on imperfect labels obtained from a traditional rule-based association method. A loss sampling mechanism is introduced to mitigate false labels. To further boost the performance, we guide the reasoning logic of AssociationNet by adding a novel ordinal loss. The proposed AssociationNet significantly outperforms the rule-based method through scene-dependent global reasoning.

Our main contributions are summarized as follows:
\begin{itemize}
    \item We proposed a scalable learning-based radar-camera fusion framework without using ground-truth labels from LiDAR, which is suitable for building a low-cost, production-ready perception system for autonomous driving applications.
    \item We designed a loss sampling mechanism to alleviate the impact of the label noise, and also invented an ordinal loss to enforce critical association logic into the model for performance enhancement.
    \item We developed a robust model via representation learning, which is capable of handling various challenging scenarios, and also outperforms the traditional rule-based algorithm by 11.6\% in terms of the F1 score.
\end{itemize}

\section{Related Work}

\subsection{Sensor Fusion}
Traditionally, different sensory modules process their data separately. A downstream sensor fusion module augments the sensory outputs (typically detected objects) to form a more comprehensive understanding of the surroundings. Such an object-level fusion method is the mainstream approach~\cite{cho2014multi, kawasaki2004standard, langer1996fusing, garcia2012data, zhong2018camera} and is still widely used on many Advanced Driver Assistance Systems (ADAS). In object-level fusion, object detection is independently performed on each sensor, and the fusion algorithm combines such object detection results to create so-called global tracks for kinematic tracking~\cite{aeberhard2012track}. 

Data association is the most critical and challenging task in object-level fusion. The precise association can easily lead to 3D object detection and multiple-object tracking solutions~\cite{aeberhard2012track, cesic2016radar}. Traditional approaches tend to manually craft various distance metrics to represent the similarities between different sensory outputs. Distance minimization~\cite{cho2014multi} and other heuristic rules are applied to find the associations. To handle the complexity and uncertainty, probabilistic models are also sometimes adopted in the association process~\cite{bar2009probabilistic}.

\subsection{Learning-Based Radar-Camera Fusion}
The learning-based radar-camera fusion algorithms can be primarily categorized into three groups, data-level fusion, feature-level fusion, and object-level fusion. The data-level fusion and feature-level fusion combine the radar and camera information at the early stage \cite{nobis2019deep, guo2018pedestrian} and the middle stage \cite{john2019rvnet, chang2020spatial, nabati2021centerfusion}, respectively, but both directly perform 3D object detection. Hence, they rely on LiDAR to provide ground-truth labels during training, which prohibits their usage to autonomous vehicles without LiDAR. 

The learning-based object-level fusion remains under-explored due to the limited information contained in the detection results. In this study, our proposed method belongs to this category in that we focus on associating radar and camera detection results. Thus, our method is more compatible with the traditional sensor fusion pipeline. On the other hand, our method also directly takes the raw camera image for further performance enhancement.

\subsection{CNN for Heterogeneous Data}
The tremendous success of CNN on structured image data inspires its application to many other types of heterogeneous data, such as sensor parameters, point clouds, and association relationships between two groups of data \cite{newell2017pixels}. In order to get compatible with CNN, a popular approach is to adapt the heterogeneous data into a form of pseudo-images. Examples include encoding camera intrinsic into images with normalized coordinates and field of view maps \cite{Facil_2019_CVPR}, projecting radar data into the image plane to form new image channels \cite{chadwick2019distant, nobis2019deep}, and the various forms of projection-based LiDAR point-cloud representations \cite{wang2020pillar, Meyer_2019_CVPR}. We adopted a similar approach in this study to handle the heterogeneous radar and camera outputs.

\subsection{Representation Learning}
Representation learning has been considered as the key to understanding complex environments and problems~\cite{bengio2013representation, lecun2015deep, kolesnikov2019revisiting}. Representation learning has been widely used in many natural language processing tasks such as word embedding \cite{mikolov2013efficient}, and many computer vision tasks, such as image classification \cite{chen2020simple}, object detection \cite{Girshick_2014_CVPR}, and keypoint matching \cite{detone2018superpoint}. In this study, we aim at learning a vector in the high-dimensional feature space as the representation for each object in the scene, in order to establish the interactions between objects as well as enable global reasoning about the scene.

\section{Problem Formulation}
We use a front-facing camera and a front-facing millimeter-wave mid-range radar for the proposed radar-camera fusion, yet the approach can be easily generalized to 360 perception with proper hardware setups. The camera intrinsic and the extrinsics of both sensors are obtained through offline calibration. The radar and camera operate asynchronously at 20Hz and 10Hz, respectively.  The field-of-views (FOVs) of the radar and camera are 120 degrees and 52 degrees, respectively. The camera is mounted under the windshield at 1.33 meters above the ground. The output of the camera sensor at each frame is an RGB image with a size of 1828 pixels (width) by 948 pixels (height), whereas the output of the radar sensor at each frame is a list of processed points with many attributes (conventionally referred to as radar pins). Since the radar used here performs internal clustering, each output radar pin is on the object level (yet the proposed fusion technique also applies to lower level detection, e.g., radar locations). There are several tens of radar pins per frame depending on the actual scene and traffic. The attributes of each radar pin are listed in Table \ref{tab:radarpin_features}. There are two noteworthy characteristics of the radar pins. First, we only consume the 2D position information in the Bird's-Eye View (BEV) without the elevation angle, due to poor resolution and large measurement noise in the elevation dimension. Second, each radar pin either corresponds to a movable object (cars, cyclists, pedestrians, etc.) or an interfering static structure such as a traffic sign, a street light, or a bridge. 
 
In this study, we focus on associating 2D bounding boxes detected from a camera image to radar pins detected in the corresponding radar frame. With precise associations, many subsequent tasks like 3D object detection and tracking become much easier if not trivial.

\begin{table}[h]
\caption{The Features of Each Radar Pin}
\label{tab:radarpin_features}
\begin{center}
\renewcommand{\arraystretch}{2} 
\begin{tabular}{llp{10cm}}
    \hline\hline
    \textbf{Feature}     &  \textbf{Explanation}\\[1pt]
    \hline
    object id  &  the id of the radar pin  \\[2pt]
    obstacle prob & \parbox{6cm}{the probability of the existence of an obstacle being detected by the radar pin} \\[2pt]
    position x & \parbox{6cm}{the x coordinate of the position of the detected obstacle in radar frame} \\[2pt]
    position y & \parbox{6cm}{the y coordinate of the position of the detected obstacle in radar frame} \\[2pt]
    velocity x & \parbox{6cm}{the velocity of the detected obstacle along the x coordinate in radar frame}   \\[2pt]
    velocity y & \parbox{6cm}{the velocity of the detected obstacle along the y coordinate in radar frame}   \\[2pt]
    \hline \hline
\end{tabular}
\end{center}
\end{table}

\begin{table}[h]
\caption{The Features of Each 2D Bounding Box}
\label{tab:bbox_features}
\begin{center}
\renewcommand{\arraystretch}{2.5} 
\begin{tabular}{llp{10cm}}
    \hline\hline
    \textbf{Feature}     &  \textbf{Explanation}\\[1pt]
    \hline
    center x  &  \parbox{6cm}{the x coordinate in the image plane of the center of the bounding box}  \\[2pt]
    center y & \parbox{6cm}{the y coordinate in the image plane of the center of the bounding box} \\[2pt]
    height & \parbox{6cm}{the height of the bounding box in the image plane} \\[2pt]
    width & \parbox{6cm}{the width of the bounding box in the image plane} \\[2pt]
    category & \parbox{6cm}{the category of the detected moving object, including \textit{sedan}, \textit{suv}, \textit{truck}, \textit{bus}, \textit{bicycle}, \textit{tricycle}, \textit{motorcycle}, \textit{person}, and \textit{unknown}}   \\[2pt]
    \hline \hline
\end{tabular}
\end{center}
\end{table}


\section{Methods}
Our proposed method mainly consists of a preprocessing step to align radar and camera data, a CNN-based deep representation learning network, AssociationNet, and a postprocessing step to extract representations and make associations. An overview of the method is shown in Fig. \ref{fig:overview} and details are explained in the following sections. 

\subsection{Radar and Camera Data Preprocessing}
Temporal and spatial alignment is performed in the preprocessing stage. For each camera frame, we look for the nearest radar frame to perform data alignment. We align the nearest radar frame to the time instant of the camera frame, by moving the radar pin locations forward/backward along the time axis under a constant velocity assumption. After the temporal alignment, the radar pins are further transformed from the radar coordinate to the camera coordinate using the known extrinsics. All the attributes of the aligned radar pins will be used in AssociationNet.

Each camera frame is first fed into a 2D object detection network to produce a list of 2D bounding boxes corresponding to the movable objects in the scene. The output attributes for each detected 2D bounding box are displayed in Table \ref{tab:bbox_features}. Though the network used in this study is an anchor-based RetinaNet ~\cite{lin2017focal} network, any 2D object detector will serve the purpose. After preprocessing, a list of temporally and spatially aligned radar pins and bounding boxes will be ready for association.

\subsection{Deep Association by Representation Learning}
We employ AssociationNet to learn a semantic representation (or a descriptor) of each radar pin and each bounding box. Under such representation, a pair of matched radar pin and bounding box will \say{look} similar, in the sense that the distance between the learned representations is small. An overview of the general process is shown in Fig.~\ref{fig:overview}. 

To leverage the powerful CNN architecture, we project each radar pin and 2D bounding box into the image plane to generate a pseudo-image, with each attribute occupying an independent channel. Specifically, each bounding box is assigned to the pixel location of its center. Each radar pin is assigned to the pixel location which is obtained through projecting its 3D location into the image plane using the camera intrinsic. The process is illustrated in \textit{Process \textbf{a}} of the Fig. \ref{fig:overview}. Next, we concatenate the raw RGB camera image with the corresponding pseudo-image to incorporate the rich pixel-level information. AssociationNet is then applied to perform representation learning. 

\begin{figure}
\begin{center}
  \includegraphics[width=1.00\linewidth]{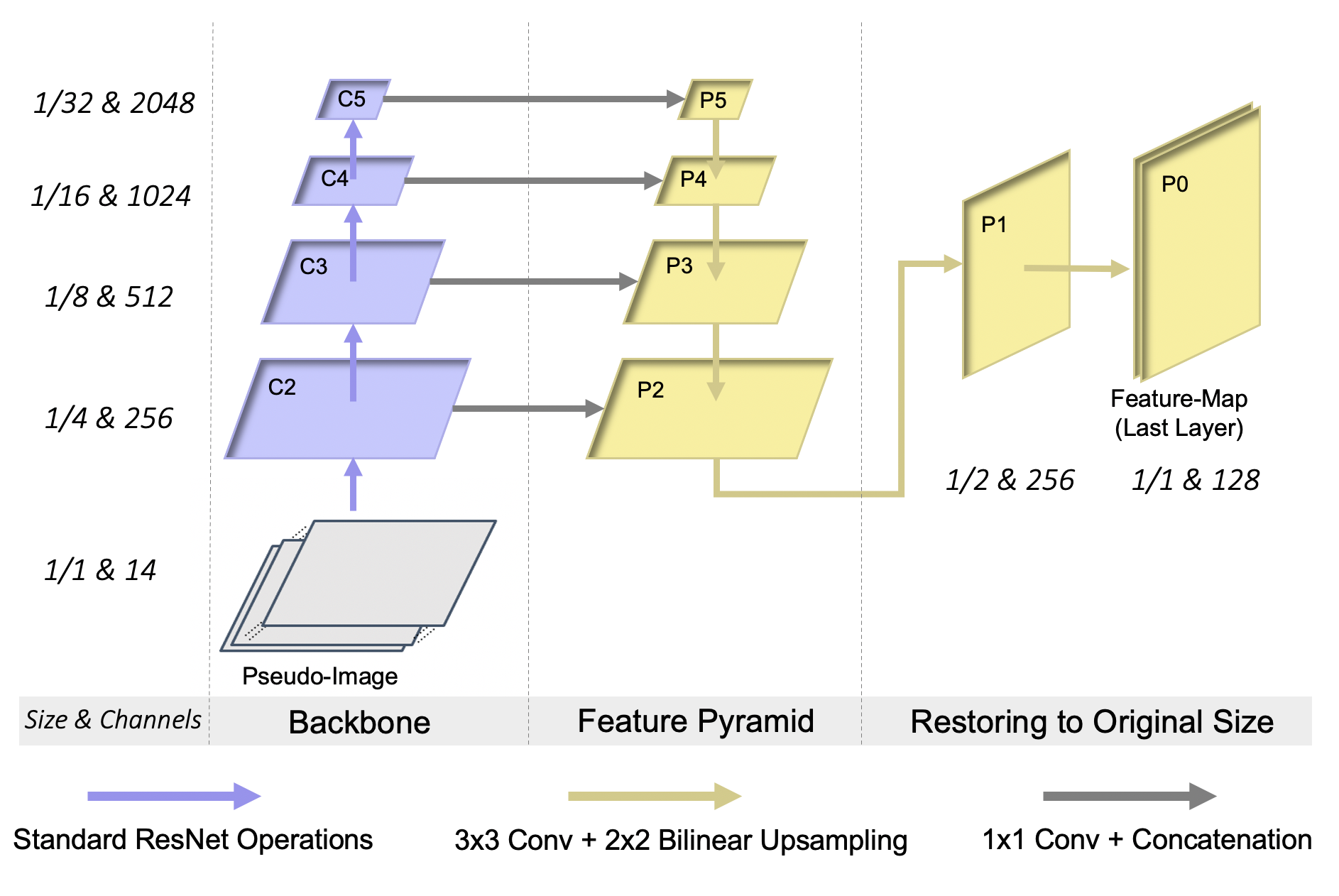}
\end{center}
  \caption{The architecture of the neural network. It consists of a ResNet-50 as the backbone, a feature pyramid for feature decoding, and two extra layers to restore the feature-map size. The feature-map in the last layer will be used to extract the representation vectors as shown in the \textit{Process \textbf{c}} of the Fig. \ref{fig:overview}}
\label{fig:network}
\end{figure}

As shown in Fig. \ref{fig:network}, the network consists of a ResNet-50 \cite{he2016deep} as the backbone, a Feature Pyramid Network \cite{lin2017feature} for feature-map decoding, and two extra layers to restore the output feature-map size to the original input size. The output feature-map contains the high-level semantic representations of radar pins and bounding boxes. As each radar pin or bounding box has a unique pixel location in the feature-map, we extract the representation vector of each of those on the output feature-map at its corresponding pixel location. The process is illustrated in \textit{Process \textbf{c}} of the Fig. \ref{fig:overview}.

The input pseudo-image contains seven radar pin channels, four bounding box channels, and three raw camera image \textit{RGB} channels. The radar pin channels include \textit{object-id}, \textit{obstacle-prob}, \textit{position-x}, \textit{position-y}, \textit{velocity-x}, \textit{velocity-y}\footnote{Positions and velocities used here are under camera coordinate, as it is after the spatial alignment step in the preprocessing.}, and a \textit{heatmap} to indicate the projected pixel location. The bounding box channels include \textit{height}, \textit{width}, \textit{category}, and also a \textit{heatmap} to indicate the pixel location. The output feature-map contains 128 channels, resulting in the dimension of the representation vector to be 64 for each radar pin and bounding box.

\begin{figure}[t]
\begin{center}
  \includegraphics[width=1.00\linewidth]{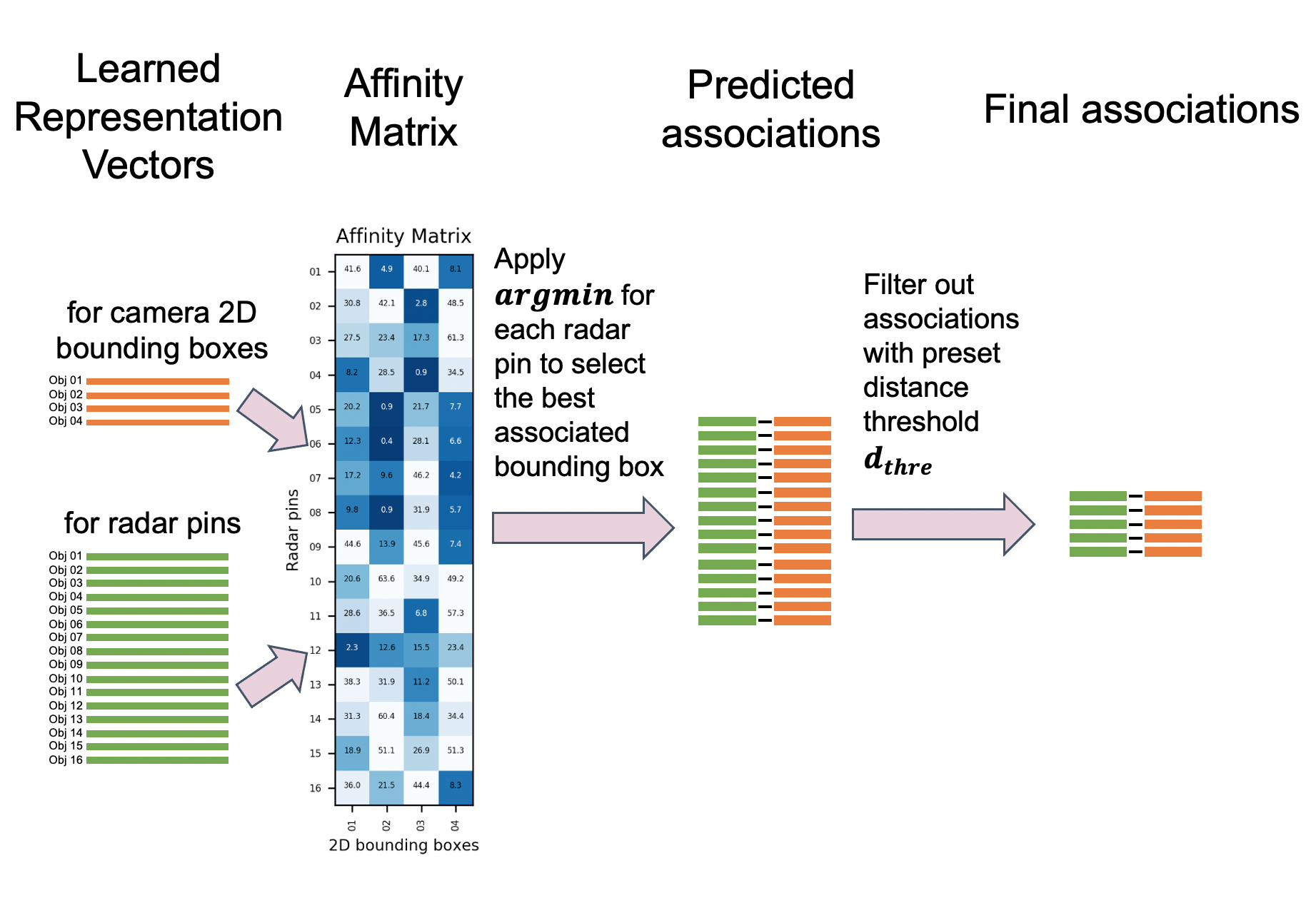}
\end{center}
  \caption{An overview of the process of obtaining final associations from the learned representation vectors.}
\label{fig:infer}
\end{figure}

\begin{figure*}[h]
\begin{center}
  \includegraphics[width=0.9\linewidth]{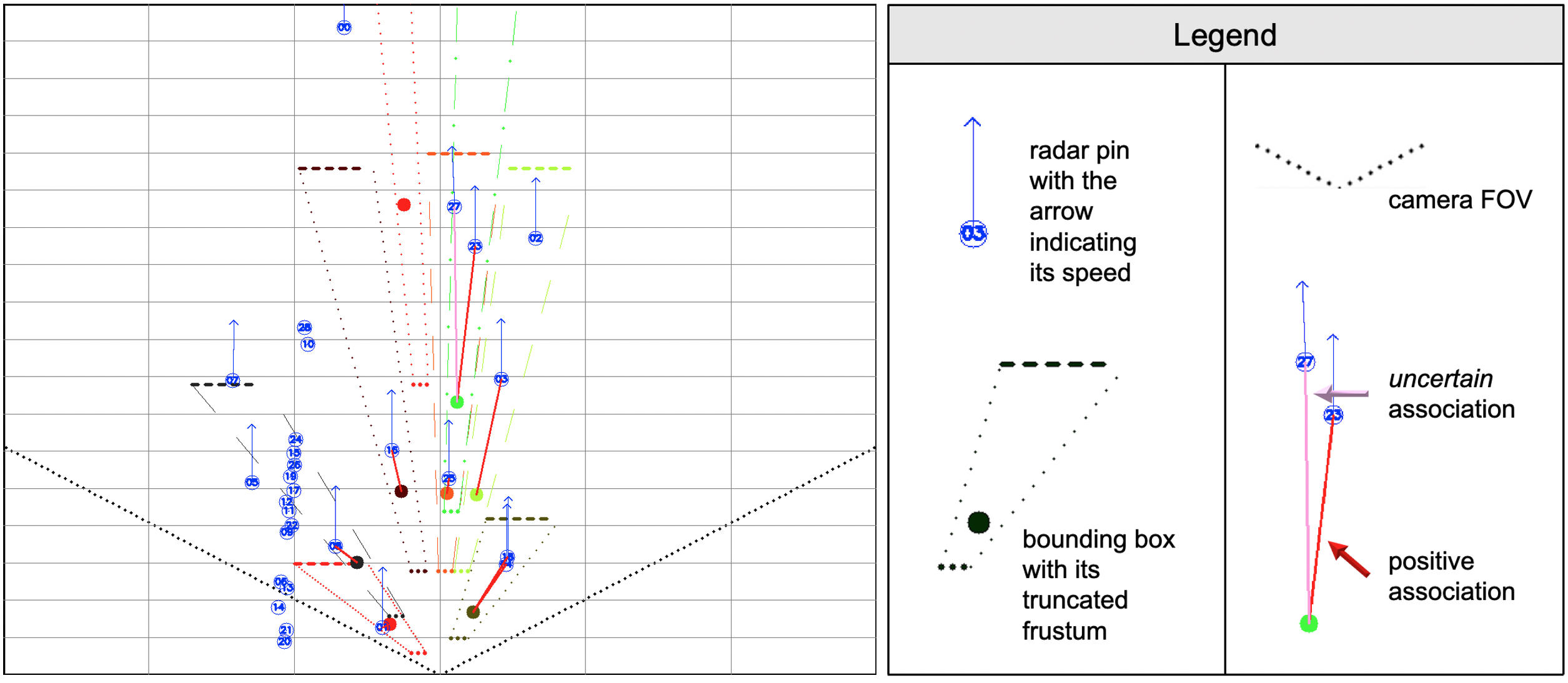}
\end{center}
  \caption[]{An illustration of radar pins, bounding boxes, and their association relationships under BEV perspective. This BEV image corresponds to the same scene as displayed in Fig. \ref{fig:showcase}. Each grid in the image represents a 10-meter-by-10-meter square in physical space. The bounding boxes are represented as solid cycles in this image. The location of a bounding box is estimated by the Inverse Projective Mapping (IPM) method from the bounding box's center, to provide a rough reference for its real 3D location. A truncated frustum accompanying each bounding box is also plotted, for better assisting human curators to determine the association relationships\footnotemark.}
\label{fig:bev_demo}
\end{figure*}

The obtained representation vector captures the semantic meaning of each radar pin and each bounding box in a high dimension space. If a radar pin and a bounding box come from the same object in the real world, we treat the pair of radar pin and bounding box as a positive sample, otherwise, it is considered as a negative sample. We try to minimize the distance between the representation vectors of any positive sample and maximize the distance between the representation vectors of any negative sample. Based on such logic, we design loss functions according to the association ground-truth labels. We pull together the representation vectors of positive samples with the following pull loss:
\begin{equation}
\label{eq:pull}
L_{pull} = \frac{1}{n_{pos}} \sum_{(i_1, i_2) \in \mathbb{POS}} \max(0, \lVert h_{i_1} - h_{i_2} \rVert - m_1)
\end{equation}
And we push apart the representation vectors of negative samples with the following push loss:
\begin{equation}
\label{eq:push}
L_{push} = \frac{1}{n_{neg}} \sum_{(i_1, i_2) \in \mathbb{NEG}} \max(0, m_2 - \lVert h_{i_1} - h_{i_2} \rVert)
\end{equation}
Here $\mathbb{POS}$ and $\mathbb{NGE}$ are the set of positive samples and the set of negative samples, respectively in each frame; $n_{pos}$ and $n_{neg}$ are the total number of associations in $\mathbb{POS}$ and $\mathbb{NGE}$ respectively;  $(i_1, i_2)$ denotes the $i^{th}$ association pair consisting of radar pin $i_1$ and bounding box $i_2$; $h_{i_1}$ and $h_{i_2}$ denotes the learned representation vectors; and $m_1$ and $m_2$ are the thresholds for the desired distances of representations among positive associations and negative associations, which were preset to be 2.0 and 8.0 in our experiments.

During inference, we calculate the Euclidean distance between the representation vectors of all possible radar-pin-bounding-box pairs. If the distance falls below a certain threshold, the radar pin and the bounding box will be considered as a successful association. More details of the inference process will be explained later.

\subsubsection{Loss Sampling}

The association labels used for supervising the learning process are ultimately from the traditional rule-based method, and hence are far from 100\% accurate. To mitigate the impact of the inaccurate labels, we first purify the labels by applying some simple filters to remove low-confidence associations, which increases the precision in the remaining association labels at the cost of the undermined recall. During the push loss calculation in the training of AssociationNet, instead of exhausting all negative pairs (a pair of a radar pin and a bounding box that is not present in the association labels), we only sample a fraction of those to be used for push loss calculation to alleviate pushing apart positive pairs by mistake. The number of sampled negative pairs is set to be equal to the number of positive ones at each frame. 

\footnotetext{The frustum is calculated also by the IPM method, from the two side edges of each bounding box. According to projective geometry, the real object detected by a bounding box has to be within the bounding box's frustum, and hence the possibly matched radar pins as well. We truncated the frustums for the ease of visualizing. The widths of each frustum at the truncated positions are set to be one meter and five meters, respectively. As the physical width of a vehicle is most likely to be within the range, the possibly matched radar pins also tend to lie within the truncated frustum.}

\subsubsection{Ordinal Loss}

One particular kind of error made by AssociationNet is that it could violate the simple ordinal rule, i.e., given two pairs of associated radar pins and bounding boxes, the farther radar pin associates to the closer bounding box. To solve this issue, an ordinal loss is introduced. 

Denote the $y$ coordinate of bounding box $i$'s bottom edge as $y_{\text{max}}^i$ and the bounding box's depth in 3D world as $d_b^i$ (which is supposedly the same as the depth of the associated radar pin $d_r^i$). For any two random bounding boxes on the same image we have the property:
\begin{equation}
\label{eq:y2d}
y_{\text{max}}^i > y_{\text{max}}^j \iff d_b^i > d_b^j
\end{equation}
The ordering of the objects in the 3D world can be interpreted as the relative vertical ordering of the bottom edges of the corresponding bounding boxes. 

Hence, we design an additional ordinal loss to enforce the self-consistency within any two associations according to the ordinal rule, which is written as:
\begin{dmath}
\label{eq:order}
L_{ord} = \frac{2}{\widehat{n_{pos}} \cdot (\widehat{n_{pos}} - 1)} \cdot \\
\sum_{\substack{
i \in \widehat{\mathbb{POS}} \\
j \in \widehat{\mathbb{POS}}
}} 
\sigma(- (d_r^{i} - d_r^{j}) \cdot (y_{\text{max}}^{i} - y_{\text{max}}^{j}) ),
\end{dmath}
where $\widehat{\mathbb{POS}}$ denotes the set of predicted positive associations and $\widehat{n_{pos}}$ is the size of the set; $i$ and $j$ are two random associations in $\widehat{\mathbb{POS}}$; $d_r^*$ represents the depth of the radar pin in an association , and $y_{\text{max}}^*$ represents the $y$ coordinate of the bounding box's bottom edge in an association; and $\sigma$ is the sigmoid function to smooth the loss values. 

Finally, the total loss is calculated as:
\begin{equation}
\label{eq:tot}
L_{tot} = L_{pull} + L_{push} + w_{ord} \cdot L_{ord},
\end{equation}
where the $w_{ord}$ is the adjustable weight to balance losses.

\subsection{Training and Inference}
The AssociationNet was trained with a batch size of 48 frames at four NVIDIA GeForce RTX 2080 Ti GPUs. The SGD optimizer was used for training at a total of 10K iterations. The learning rate was set to be $10^{-4} $ initially, and then was decreased by a factor of 10 at the end of 8K iterations and 9K iterations, respectively. 

At the inference time, the representation vectors for all radar pins and bounding boxes are first predicted using the trained model. An affinity matrix is then calculated, where each matrix element corresponds to the distance between the representations of a radar pin and a bounding box. In reality, each bounding box may be associated with multiple radar pins (this is usually the case where the corresponding vehicles are of large sizes, such as trailer trucks and buses.), while each radar pin can only match to at most one bounding box. As a result, we associate each radar pin to the bounding box with the smallest distance in the affinity matrix. Lastly, the improbable associations with a distance larger than a threshold are filtered out, which usually consists of radar pins from interfering static objects. The whole inference process is summarized in Fig.~\ref{fig:infer}.

\subsection{Evaluation}
The predicted associations are compared against human-annotated ground-truth associations in the test dataset. We use precision, recall, and F1 score as the metrics for evaluating the performance. 

In some very complicated scenes, correctly associating all radar pins and bounding boxes is very challenging even for human annotators. Therefore, we mark those plausible but dubious associations as \say{\textit{uncertain}} in the evaluation process. An example is shown in Fig. \ref{fig:bev_demo}. For those \say{\textit{uncertain}} associations, they are counted as neither positive nor negative associations, which will be excluded from both true and false positive predictions. 

\begin{figure*}[h]
    \centering
      \begin{subfigure}[b]{1\textwidth}
          \centering
          \includegraphics[width=1\linewidth]{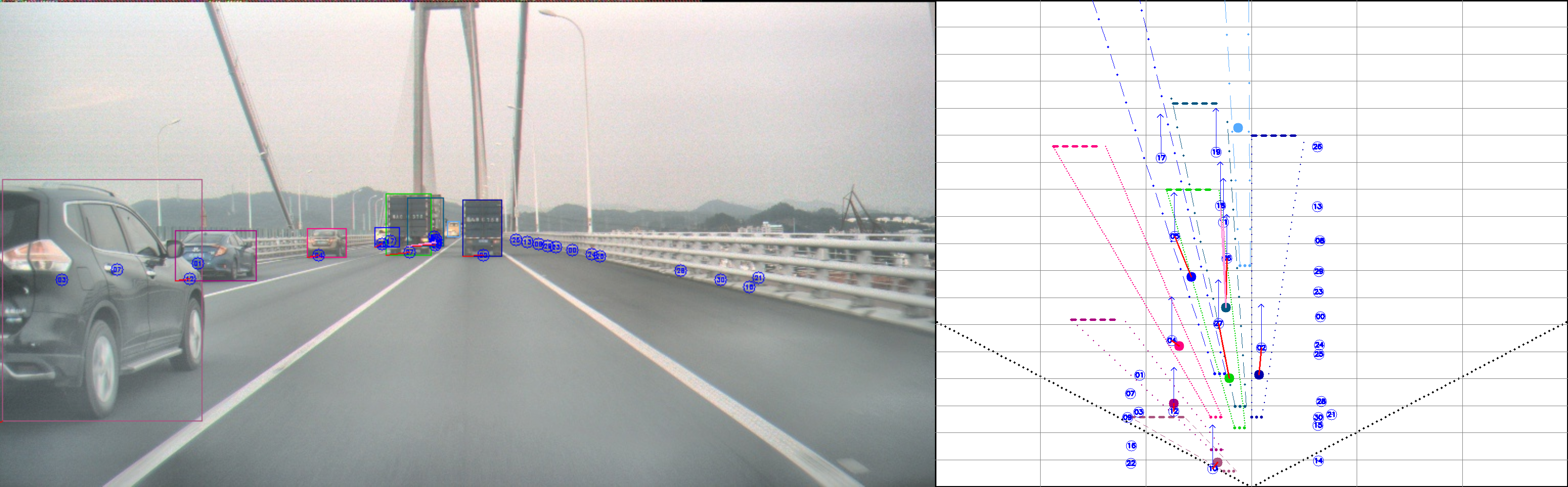}
      \end{subfigure}
      \begin{subfigure}[b]{1\textwidth}
          \centering
          \includegraphics[width=1\linewidth]{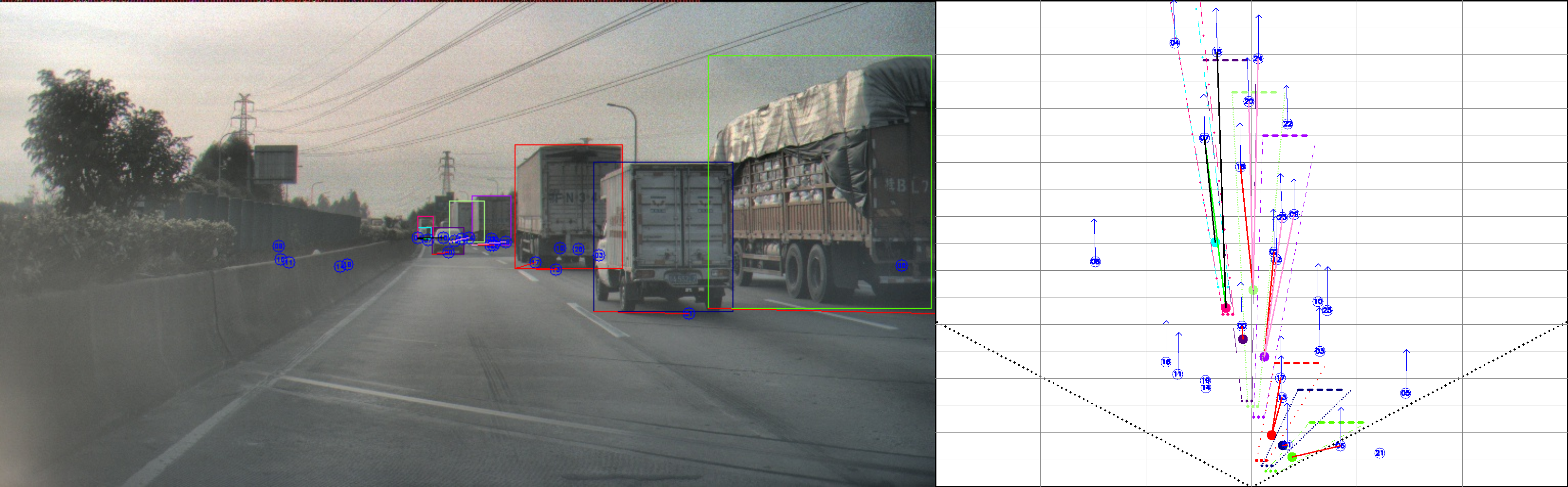}
      \end{subfigure}
\caption{Examples of AssociationNet predictions. Here, the red solid lines represent the true-positive associations; and the pink solid lines represent predicted positive associations but labeled as \textit{uncertain} in the ground-truth. In the second example, the added green lines represent the false-positive predictions; and the added black lines represent the false-negative predictions.
Also, note that each bounding box on the left corresponds to a solid circle with the same color on the right.
}
\label{fig:result_demo}
\end{figure*}

\section{Experiments and Discussion}

\subsection{Dataset}
The AssociationNet was trained and evaluated on an in-house dataset with 12 driving sequences collected by a testing fleet, which consists of 14.8 hours of driving in various driving scenarios, including highway, urban, and city roads. The radar and camera were synchronized at 10 Hz initially and further downsampled to 2 Hz, in order to reduce the temporal correlation among adjacent frames. Eleven sequences out of the twelve were used for training with the other one left for the test. Therefore, there are 104,314 synchronized radar and camera frames in the training dataset, and 2,714 in the test dataset. For the training data, the association labels were generated by a traditional rule-based algorithm with additional filtering to increase the precision. For the test data, we manually curated the labels with human annotators to obtain high-quality ground-truth labels.

\subsection{Effect of Loss Sampling}
We studied the effect of loss sampling on the AssociationNet's performance. Experiments were conducted with \textit{no sampling} (meaning that all the negative pairs present in the label are used for push loss calculation), and loss sampling with different sampling ratios. The sampling ratio is defined as the ratio between the number of positive pairs and the number of negative pairs at each frame. The result is shown in Table \ref{tab:effect_sample}. We can see that the best sampling ratio is 1:1 with the loss sampling mechanism, which boosts the performance by 1.1\% in terms of the F1 score.

\begin{table}[]
\caption{The Effect of Loss Sampling}
\label{tab:effect_sample}
\begin{center}
\begin{tabular}{c|c}
\hline\hline
\textbf{Sample Ratio}     &  \textbf{Performance} \\[1pt]
\textbf{ }     &  Precision / Recall / F1 \\[1pt]
\hline
\textit{no sampling} & 0.896 / 0.925 / 0.911 \\[1pt]
1:2 & 0.901 / 0.931 / 0.916 \\[1pt]
1:1 & 0.906 / 0.939 / 0.922 \\[1pt]
2:1 & 0.899 / 0.933 / 0.915 \\[1pt]
3:1 & 0.899 / 0.929 / 0.914 \\[1pt]
\hline\hline
\end{tabular}
\end{center}
\end{table}

\subsection{Effect of Ordinal Loss}
The effect of the ordinal loss is shown in Table \ref{tab:effect_ord}. The ordinal loss can facilitate both precision and recall to some degree. With the optimal loss weight, the performance is boosted by 1.8\% in terms of the F1 score.

\begin{table}[]
\caption{The Effect of Ordinal Loss}
\label{tab:effect_ord}
\begin{center}
\begin{tabular}{c|c}
\hline\hline
\textbf{Loss Weight}     &  \textbf{Performance} \\[1pt]
$w_{ord}$     &  Precision / Recall / F1 \\[1pt]
\hline
0.0 & 0.897 / 0.912 / 0.904 \\[1pt]
0.5 & 0.897 / 0.923 / 0.910 \\[1pt]
1.0 & 0.899 / 0.931 / 0.915 \\[1pt]
2.0 & 0.906 / 0.939 / 0.922 \\[1pt]
5.0 & 0.889 / 0.918 / 0.903 \\[1pt]
\hline\hline
\end{tabular}
\end{center}
\end{table}

\subsection{Comparison with Rule-Based Algorithm}
We compared the performance of AssociationNet with the traditional rule-based algorithm, as shown in Table \ref{tab:compare}. Notably, though the traditional rule-based algorithm was used to generate association labels to supervise the training of AssociationNet, AssociationNet significantly outperforms the rule-based alternative. This demonstrates the inherent robustness of learning-based algorithms in handling complex scenarios.

\begin{table}[]
\caption{Comparison with Rule-based Algorithm}
\label{tab:compare}
\begin{center}
\begin{tabular}{c|c}
\hline\hline
\textbf{Algorithm}  &  \textbf{Performance} \\[1pt]
\textbf{ }     &  Precision / Recall / F1 \\[1pt]
\hline
Rule-based & 0.890 / 0.736 / 0.806 \\[1pt]
Learning-based & 0.906 / 0.939 / 0.922 \\[1pt]
\hline\hline
\end{tabular}
\end{center}
\end{table}

\subsection{Visualization}
Examples of the predicted associations are shown in Fig. \ref{fig:result_demo}. Despite multiple big trucks present in both examples, AssociationNet correctly predicted their associations, which demonstrates the robustness of the algorithm. On the other hand, in the second example, there are two bounding boxes incorrectly associated, with one bounding box having no predicted associations and the other associated to a wrong radar pin. The two bounding boxes correspond to vehicles at the very far range. The mistakes are largely due to the small sizes of the objects in the camera image and also the heavy occlusions.

\section{Conclusion}
In this work, we developed a scalable learning-based radar-camera fusion algorithm, without using LiDAR for ground-truth labels generation. Such a solution has many practical merits at the current technological stage, including low cost, low maintenance, high reliability, and more importantly, readiness for mass production. We employed deep representation learning to tackle the challenging association problem, with the benefits of enabled feature-level interaction and global reasoning. We also designed a loss sampling mechanism and a novel ordinal loss to mitigate the impact of label noise and enforce critical human logic into the learning process. Although imperfect labels generated by a traditional rule-based algorithm were used to train the network, our proposed algorithm outperforms the rule-based teacher by 11.6\% in terms of the F1 score.

\section{Acknowledgements}
We sincerely thank Xinzhou Wu and Tao Wang for insightful discussions, and thank Yongbo Tan and Yunfei Cheng for labeling supports.

{\small
\bibliographystyle{ieee_fullname}
\bibliography{egbib}
}

\end{document}